# Benchmarking Cross-Lingual Semantic Alignment in Multilingual Embeddings

Wen G. Gong
wen.gong.research@gmail.com

## Abstract

With hundreds of multilingual embedding models available, practitioners lack clear guidance on which provide genuine cross-lingual semantic alignment versus task performance through language-specific patterns. Current task-driven benchmarks (MTEB) may mask fundamental alignment shortcomings. We introduce Semantic Affinity ($SA$), a bounded [0,1] metric measuring inter-lingual to intra-lingual spread ratio using cosine distance, combined with PHATE visualization in our Semanscope framework. Benchmarking 13 models across 4 datasets (52 experiments) reveals a three-tier structure: (1) Top BERT models (LaBSE $SA = 0.70$, USE $SA = 0.68$, S-BERT $SA = 0.68$) achieve strong alignment via translation-pair supervision; (2) LLM embeddings plateau at $SA \approx$ 0.55-0.61 regardless of 0.6B→8B scale; (3) MLM-only BERT models (mBERT, XLM-R, $SA < 0.50$) fail despite 100+ language training. Training objective, not architecture or scale, determines alignment. Oracle Bone primitives (1200 BCE) expose semantic drift —models learn corpus patterns rather than cognitive primitives. This work provides semantic benchmarking to help practitioners select quality multilingual embeddings from hundreds of available models, showing cross-lingual alignment requires explicit translation supervision, not merely model scale or multilingual data.

## 1. Introduction

Practitioners face hundreds of multilingual embedding models—Hugging Face hosts over 500 multilingual sentence transformers, OpenRouter provides API access to dozens of LLM embeddings—yet lack clear guidance on which models provide genuine cross-lingual semantic alignment. Current evaluation focuses on downstream task performance (MTEB, XTREME), but strong task performance may mask poor cross-lingual semantic alignment. Recent multilingual embedding methods rely on the assumption that "a translated sentence should be mapped to the same location in the vector space as the original sentence" (Reimers & Gurevych, 2020). However, this foundational assumption has never been rigorously tested across diverse models (both BERT and LLM), training objectives, and semantic domains.

We introduce Semanscope, combining PHATE manifold learning with Semantic Affinity metric to systematically test the translation-equivalence assumption. $SA$ measures the ratio of inter-lingual to intra-lingual spread using cosine distance, providing a bounded [0,1] metric where higher values indicate stronger cross-lingual alignment. This absolute scale enables practitioners to quickly filter models: $SA \geq 0.60$ indicates strong alignment suitable for cross-lingual tasks, while $SA < 0.50$ reveals alignment failure despite potentially strong task performance. PHATE visualization with integrated $SA$ legends enables both quantitative measurement and qualitatively visual validation in single-figure diagnostics.

Benchmarking 13 models across 4 datasets (52 experiments) reveals that alignment is training-dependent, not architecture-dependent or scale-dependent. Only models with explicit translation-pair supervision or semantic similarity objectives achieve strong alignment ($SA \geq 0.60$). This work provides semantic benchmarking to help practitioners select quality multilingual embeddings from hundreds of available models, complementing task-based evaluation and identifying critical requirements for cross-lingual semantic alignment.

## 2. Related Work

Multilingual Embedding Assumptions: Reimers & Gurevych (2020) assume "a translated sentence should be mapped to the same location in the vector space as the original sentence" for their knowledge distillation approach. LaBSE (Feng et al., 2022) and Universal Sentence Encoder (Cer et al., 2018; Yang et al., 2020) use similar assumptions with translation-pair supervision or semantic similarity objectives. Recent work on cross-lingual transferability (Artetxe et al., 2020; Vulić et al., 2020) explores alignment quality but lacks unified metrics. Bilingual sentence embeddings (Guo et al., 2018) focus on parallel corpus mining rather than semantic alignment assessment. However, no prior work has systematically tested whether this assumption holds across different model architectures (from BERT (Devlin et al., 2019) to LLMs), training objectives, semantic domains, and language pairs. Semanscope provides not only the first rigorous empirical test, but also discriminative capability in benchmarking embedding models in terms of semantic alignment.

Benchmarking Gap: MTEB (Muennighoff et al., 2022) and XTREME (Hu et al., 2020) evaluate task performance but tolerate language clustering—retrieval succeeds via intra-lingual patterns even when languages separate. This leaves practitioners unable to distinguish models with genuine cross-lingual alignment from those achieving task performance through language-specific shortcuts. MLM-only models like mBERT (Devlin et al., 2019) and XLM-R (Conneau et al., 2020) achieve strong monolingual performance despite poor cross-lingual alignment. Recent LLM embeddings (Neelakantan et al., 2022) and contrastive pre-training approaches (Wang et al., 2022) lack systematic alignment evaluation. Our $SA$ metric directly measures whether models satisfy the translation-equivalence assumption by quantifying inter-lingual vs intra-lingual spread, providing actionable guidance for model selection.

PHATE Visualization: PHATE (Moon et al., 2019) preserves both local and global manifold structure via diffusion-based distance computation, complementing other dimensionality reduction techniques like t-SNE (van der Maaten & Hinton, 2008) and UMAP (McInnes et al., 2018). We extend PHATE with automatic $SA$ metric calculation, creating diagnostic charts combining visual validation with quantitative assessment. Prior semantic similarity benchmarks (Agirre et al., 2012) evaluate monolingual alignment without cross-lingual extension.

## 3. Semantic Affinity Metric Definition

Given M words translated across L languages, let $\mathrm{E}_\ell \in \mathbb{R}^{M \times d}$ be the embedding matrix for language $\ell$.

Intra-Lingual Spread measures semantic diversity within each language, providing normalization baseline. We use cosine distance as it is invariant to embedding magnitude:

$$D_{\mathrm{intra}} = \frac{1}{L} \sum_{\ell=1}^{L} \frac{1}{N_\ell} \sum_{i<j} \left( 1 - \frac{e_i^{(\ell)} \cdot e_j^{(\ell)}}{\|e_i^{(\ell)}\| \|e_j^{(\ell)}\|} \right)$$

where $N_\ell$ = number of unique word-pairs in language $\ell$.

Inter-Lingual Spread measures cosine distances of translation equivalents between languages:

$$D_{\mathrm{inter}} = \frac{1}{M'} \sum_{w=1}^{M'} \frac{1}{K} \sum_{i<j} \left( 1 - \frac{e_w^{(i)} \cdot e_w^{(j)}}{\|e_w^{(i)}\| \|e_w^{(j)}\|} \right)$$

where K = L(L-1)/2 language pairs, M' = number of expanded word pairs.

Semantic Affinity defines a ratio that is normalized and bounded to (0, 1]:

$$SA = \frac{D_{\mathrm{intra}}}{D_{\mathrm{intra}} + D_{\mathrm{inter}}}$$

Interpretation:

- $SA \geq 0.60$: Great alignment (Tier 1: translation-pair supervision)

- $0.50 \leq SA < 0.60$: Good alignment (Tier 2: LLM embeddings, task-dependent)

- $SA < 0.50$: Non-alignment (Tier 3: language separation, alignment failure)

Why Cosine Distance? Cosine distance is magnitude-invariant, making $SA$ scores comparable across datasets with different embedding scales. Euclidean distance is scale-dependent—models show inconsistent $SA$ scores across datasets due to varying absolute magnitudes rather than true alignment quality. See Appendix A.5 for empirical evidence and detailed comparison. Nevertheless, we provide Euclidean distance $SA$ in Appendix A.1 for reference.

## 4. Experimental Setup

### 4.1. Datasets

We benchmark on four complementary datasets (952 total words) spanning modern concepts (DS1: Gärdenfors conceptual spaces, 349 words), cultural nuance (DS2: kinship, philosophical concepts, 190 words), linguistic edge cases (DS3: untranslatables, polysemes, false friends, 86 words), and ancient primitives (DS4:

Oracle Bone Script 1200 BCE, 327 words). See Table C.1 in Appendix C.1 for complete dataset specifications. All datasets use English-Chinese translation pairs will be released publicly upon publication.

Domain-Specific Evaluation Guidance: These four datasets serve research and illustrative purposes, demonstrating Semanscope's methodology across diverse semantic domains. Practitioners should create domain-specific datasets tailored to their unique use cases for actionable model selection. For example: life sciences researchers should benchmark models on specialized terminology (gene names, protein functions, anatomical terms); financial institutions should test domain-relevant concepts (risk metrics, market instruments, regulatory terms); legal applications require evaluation on jurisdiction-specific terminology. Models showing $SA \geq 0.60$ on general benchmarks may exhibit different alignment quality on specialized vocabularies—domain-specific evaluation ensures the selected model performs well on the actual deployment vocabulary. Semanscope's framework enables practitioners to run custom benchmarks with their own translation pairs, providing use-case-specific model rankings rather than relying solely on general-purpose benchmark results.

## 4.2. Models

We evaluate 13 state-of-the-art multilingual embedding models spanning BERT architectures (LaBSE, mBERT, XLM-R, Sentence-BERT, USE, E5-Large) and LLM-based embeddings (OpenAI text-embedding-3 series, Gemini-001, Qwen3 series). Models vary in training objectives (translation pairs, MLM, next-token prediction, instruction tuning), parameter scales (110M-8B), and architectural approaches. See Appendix C.2 for complete model specifications.

## 4.3. PHATE Visualization

PHATE (Potential of Heat-diffusion for Affinity-based Transition Embedding) preserves both local neighborhoods and global manifold structure, making it ideal for semantic alignment validation. The basic PHATE configurations are: (knn=15, decay=40, t='auto', dimensions=2). StandardScaler normalization, duplicate filtering. See Appendix B for complete configuration details.

With PHATE chart, (1) Visual ground truth—interleaved clusters indicate low inter-lingual spread; (2) Failure diagnostics—identifies which language pairs fail; (3) Manifold structure preservation enables evaluation of universal semantic structure; (4) Integrated $SA$ metric confirms visual pattern quantitatively.

## 4.4. Semanscope Framework

Semanscope is our purpose-built benchmark tool combining $SA$ metric computation with PHATE visualization. Its key features are:

1. Automatic $SA$ Legend Generation: Computes $SA$ metrics (Cosine primary, Euclidean supplementary) and overlays them on PHATE plots (top-right corner), creating publication-ready single-figure diagnostics

2. Per-Word Embedding Cache: 30× efficiency gain by caching at word level (327 unique embeddings vs 9,397 phrase combinations)

3. True Multiprocessing: 4.6× speedup using loky backend (multi-core CPU), avoiding Python GIL limitations

4. Dual Metrics: Reports both Cosine (primary) and Euclidean (supplementary) distance $SA$ scores for comprehensive evaluation

## 5. Results

### 5.1. Three-Tier Structure of Model Performance

Figure 1 shows average $SA$ scores (cosine) across 4 datasets for 13 models (see Tables D.1-D.2 in Appendix D for complete results). We observe a three-tier structure that provides clear guidance for practitioners choosing among hundreds of available models: (1) Top BERT models (LaBSE $SA = 0.70$, USE $SA = 0.68$, Sentence-BERT $SA = 0.68$) achieve $SA \geq 0.60$ (great alignment) via translation-pair supervision or semantic similarity objectives—deploy with confidence for cross-lingual tasks; (2) LLM embeddings plateau at $0.50 \leq SA \leq 0.61$ (good alignment, task-dependent) regardless of 0.6B→4B→8B scale—suitable for specific use cases but verify alignment quality; (3) MLM-only BERT models (mBERT $SA = 0.50$, XLM-R $SA = 0.45$) fail at $SA < 0.50$ (non-alignment) despite 100+ language training—avoid for cross-lingual applications despite strong task performance. This reveals translation-pair supervision or semantic similarity objectives are essential—multilingual pre-training alone is fundamentally insufficient.

### 5.2. Dataset Difficulty Gradient Reveals Model Robustness

Figure 2 shows the complete heatmap of all 13 models across 4 datasets, revealing both the three-tier structure and the dataset difficulty gradient. Color intensity (red→low $SA$, yellow→medium, green→high) clearly shows DS1/DS2 as easier and DS3/DS4 as harder.

We observe that (1) Top BERT models excel on modern concepts but show steeper drops (-25-31%) on edge cases/ancient text; (2) LLM embeddings more robust across diverse domains (smaller drop, -15-17%), suggesting broader training distribution; (3) Failed models consistently show poor performance across all datasets.

Additionally, ZiNets Oracle Bone primitives (DS4, 769 words) serve as stress test—even universal concepts humans recognized for 3000 years challenge modern embeddings due to semantic drift. Top BERT models' steeper drop suggests they excel on training-distribution-proximate data but struggle with out-of-distribution ancient/edge cases.

With E5-Large model, instruction tuning optimizes task performance (MTEB), not primitive concept alignment. For Qwen3 models with scaling ($0.6B \rightarrow 4B \rightarrow 8B$), next-token prediction provides weak cross-lingual binding.

### 5.3. Visual Evidence by PHATE

Figure 3 shows 4 representative PHATE plots ($2 \times 2$ grid) with automatic $SA$ legends (cosine primary), all using DS1 (Peterg, 349 words) for fair comparison. More visualization charts are provided in Appendix E.

We observe that (1) LaBSE (Top-left, $SA = 0.807$) has beautiful language interleaving with translation pairs co-locating tightly; (2) OpenAI-3-Large (Top-right, $SA = 0.644$) shows moderate cross-lingual mixing with visible clustering; (3) Qwen3-0.6B (Bottom-left, $SA = 0.490$) exhibits failure—partial language separation at threshold; (4) mBERT (Bottom-right, $SA = 0.507$) exhibits significant language separation despite 104-language MLM training.

## 6. Discussion

### 6.1. Testing the Translation-Equivalence Assumption

Reimers & Gurevych (2020) built their multilingual approach on the assumption that "a translated sentence should be mapped to the same location in the vector space as the original sentence." Our 52 experiments show this assumption is training-dependent, not architecture-dependent. Translation-pair supervision (LaBSE $SA = 0.70$) and semantic similarity objectives (USE $SA = 0.68$, S-BERT $SA = 0.68$) enforce alignment explicitly. Next-token prediction (LLMs $SA \approx 0.55$-$0.61$) optimizes sequential coher-ence without cross-lingual binding—parameter scaling from 0.6B to 8B shows diminishing returns. MLM-only training (mBERT $SA = 0.50$, XLM-R $SA = 0.45$) learns language-specific patterns despite 100+ language coverage.

Instruction tuning creates a gap between task optimization and semantic grounding. E5-Large achieves $SA = 0.55$ despite BERT architecture and 335M parameters—models embed instructions/queries effectively while failing to bind translation variants. This reveals that multilingual data, large scale, and instruction tuning are insufficient alone. Cross-lingual alignment requires explicit alignment objectives during pre-training.

### 6.2. Semantic Drift in Oracle Bone Primitives

Chinese characters inscribed on Oracle Bone (e.g., 日 sun, 水 water) represent universal human experiences from 1200 BCE, yet models show worse alignment on DS4 (ancient primitives) than DS2 (cultural nuance). Over three millennia, These 2 characters accumulated diverse semantic extensions: 生日 birthday, 日记 diary, 日常 routine/daily, 日出 sunrise, 日落 sunset, 节日 holiday, 日程 schedule/itinerary, 日本 Japan, 水平 level, 水果 fruit, 香水 perfume, 水手 sailor, 水稻 rice, 泪水 tear, 淡水 fresh-water, 水库 reservoir. Models trained on contemporary corpora capture extended semantics, not the original primitive—they learn corpus co-occurrence patterns rather than cognitive primitives. This exposes a fundamental limitation: embedding models assign fixed vectors to context-dependent meanings.

### 6.3. Implications for Model Selection

Task benchmarks (MTEB, XTREME) tolerate language clustering—retrieval succeeds via intra-lingual patterns even when languages separate. mBERT achieves reasonable MTEB scores while failing cross-lingual alignment ($SA < 0.50$). This creates a model selection challenge: practitioners cannot rely on task performance alone to identify models with genuine cross-lingual semantic alignment.

SA + PHATE directly measure whether models satisfy the translation-equivalence assumption, providing actionable guidance: (1) For applications requiring genuine cross-lingual understanding (multilingual search, cross-cultural moderation, zero-shot transfer), choose models with $SA \geq 0.60$ (LaBSE, USE, S-BERT); (2) For domain-specific tasks where alignment matters less, LLM embeddings ($0.50 \leq SA \leq 0.61$) may suffice; (3) Avoid MLM-only models ($SA < 0.50$) for cross-lingual applications despite their strong monolingual performance. This bounded [0,1] scale enables practitioners to quickly filter hundreds of available models based on alignment requirements, complementing task-based evaluation with semantic quality assessment.

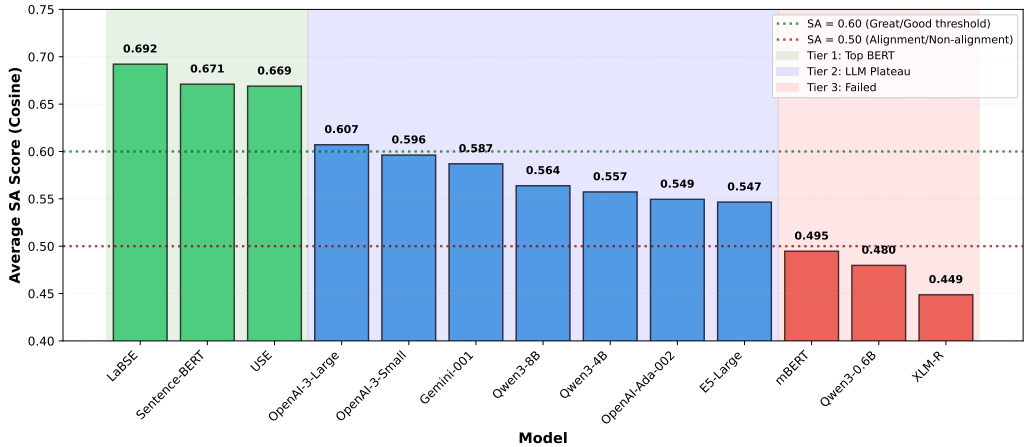

Figure 1. Three-Tier Performance Structure (Average $SA$ cosine across 4 datasets)

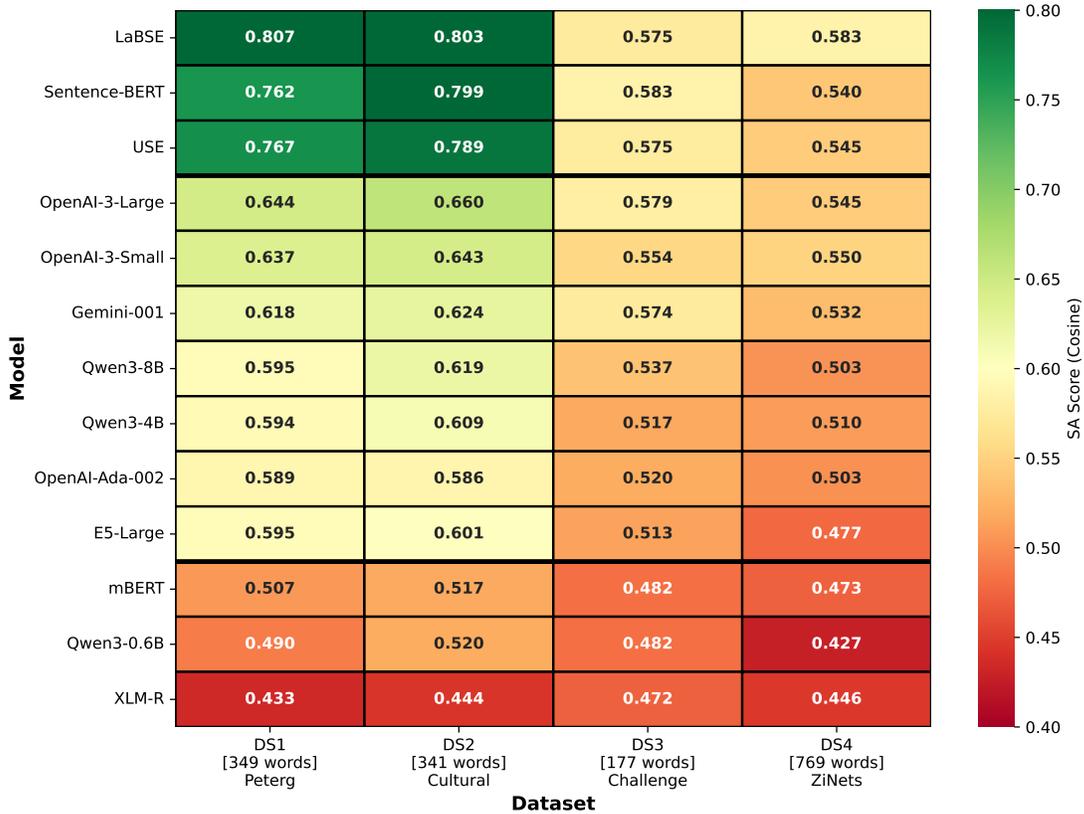

Figure 2. Dataset Difficulty Gradient Heatmap (13 Models × 4 Datasets, $SA$ cosine)

## 7. Limitations

This study has the following limitations: (1) We evaluated English-Chinese translation pairs; $SA$ scores may

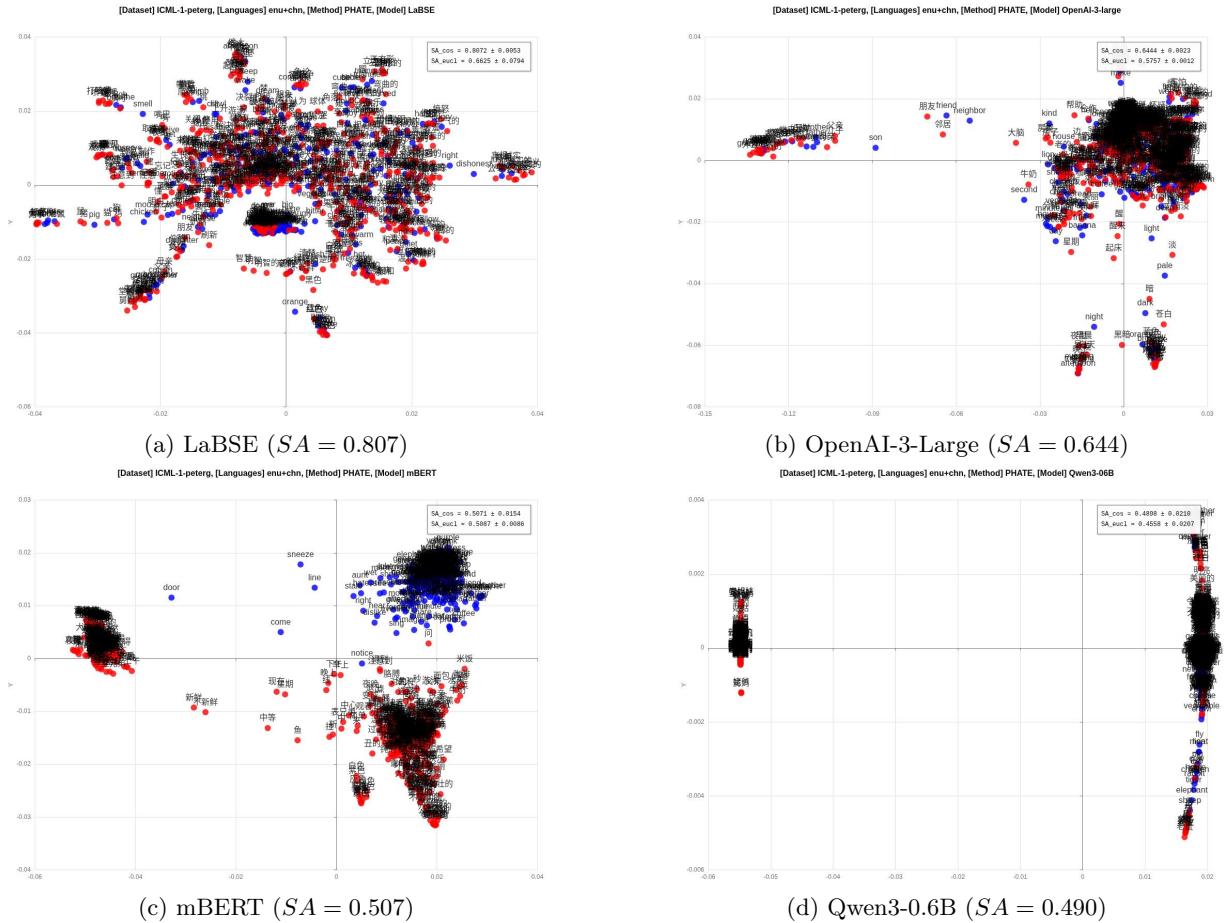

(a) LaBSE ($SA = 0.807$)

(b) OpenAI-3-Large ($SA = 0.644$)

(c) mBERT ($SA = 0.507$)

(d) Qwen3-0.6B ($SA = 0.490$)

Figure 3. PHATE Manifolds - 2×2 Grid (DS1: Peterg, 349 words, $SA$ cosine)

differ for other language pairs, particularly those with greater linguistic distance (e.g., English-Arabic). (2) PHATE hyperparameter sensitivity (knn, decay, t) may affect visual alignment assessment, though $SA$ metric computation remains hyperparameter-free. (3) while our 4 datasets span diverse semantic domains, practitioners should construct domain-specific benchmarks for their actual deployment vocabulary, as alignment quality may vary between general concepts and specialized terminology. (4) $SA$ measures static alignment without capturing dynamic context-dependent semantics—future work should explore contextualized alignment metrics for polysemous words.

## 8. Conclusion

With hundreds of multilingual embedding models available, practitioners need clear guidance on which provide genuine cross-lingual semantic alignment. We introduce Semanscope, combining PHATE manifold learning with Semantic Affinity ($SA$) metric using co-sine distance to test the translation-equivalence assumption underlying multilingual embeddings. Benchmarking 13 models across 4 datasets (52 experiments) reveals a three-tier structure that enables actionable model selection: (1) Top BERT models (LaBSE $SA = 0.70$, USE $SA = 0.68$, S-BERT $SA = 0.68$) achieve strong alignment via translation-pair supervision—deploy with confidence; (2) LLM embeddings plateau at $SA \approx 0.55$-$0.61$ regardless of $0.6B \rightarrow 8B$ scale—suitable for specific use cases; (3) MLM-only BERT models (mBERT $SA = 0.50$, XLM-R $SA = 0.45$) fail despite 100+ language training—avoid for cross-lingual applications.

Architecture comparison shows training objective matters more than structure: LaBSE and mBERT share BERT-base architecture but differ by +40% $SA$ (0.70 vs 0.50) due to training objective alone. Oracle Bone primitives (DS4) expose semantic drift—universal concepts from 1200 BCE challenge models more than cultural nuance, revealing models learn corpus patterns rather than cognitive primitives.

This work provides semantic benchmarking to help practitioners select quality multilingual embeddings from hundreds of available models, complementing task-based evaluation (MTEB). $SA$'s bounded $[0,1]$ scale with magnitude-invariant cosine distance enables practitioners to quickly filter models: $SA \geq 0.60$ indicates strong alignment, $SA < 0.50$ reveals alignment failure despite potentially strong task performance. PHATE with integrated legends provides single-figure diagnostic charts. The finding that explicit translation supervision is essential—not merely model scale, multilingual data, or instruction tuning—informs both model selection and future multilingual architecture design.

## Acknowledgments

This research represents a novel Human-AI collaboration where domain expertise, research vision, and execution are amplified through AI assistance. The author provided research direction, domain knowledge (Oracle Bone Script, Chinese linguistics), and quality control, while AI assistants contributed dataset design, coding, visualization enhancements (automatic SA legend integration), and analytical review.

The author would like to thank Albert W. Gong for valuable feedback on semantic affinity definition. The author acknowledges AI assistance from Claude (Anthropic) and Gemini (Google) for literature research, coding, data analysis, and manuscript preparation.

Dataset attributions: ICML-1-peterg based on Gärdenfors (2014); ICML-2-cultural-nuance and ICML-3-challenge proposed by Claude Sonnet 4.5 (Anthropic); ICML-4-zinets developed by the author from Oracle Bone Script primitives.

## Appendix A: Mathematical Formulation

### A.1 Euclidean Distance (Supplementary)

For reference, we also compute $SA$ using Euclidean distance, though it is scale-dependent and less reliable across datasets.

Intra-Lingual Spread (Euclidean):

$$D_{\text{intra}}^{\text{eucl}} = \frac{1}{L} \sum_{\ell=1}^{L} \sqrt{\frac{1}{N_\ell} \sum_{i<j} \|e_i^{(\ell)} - e_j^{(\ell)}\|^2}$$

Inter-Lingual Spread (Euclidean):

$$D_{\text{inter}}^{\text{eucl}} = \frac{1}{M'} \sum_{w=1}^{M'} \sqrt{\frac{1}{K} \sum_{i<j} \|e_w^{(i)} - e_w^{(j)}\|^2}$$

Semantic Affinity (Euclidean):

$$SA_{\text{eucl}} = \frac{D_{\text{intra}}^{\text{eucl}}}{D_{\text{intra}}^{\text{eucl}} + D_{\text{inter}}^{\text{eucl}}}$$

### A.2 Why RMS for Euclidean, Mean for Cosine

Euclidean: Uses Root Mean Square (RMS) because we compute squared distances $\|e_i - e_j\|^2$, making RMS the natural L2 norm:

$$\text{RMS} = \sqrt{\frac{1}{N} \sum d_i^2}$$

Cosine: Uses simple mean because cosine distance is already well-defined in [0,2]:

$$\text{Mean} = \frac{1}{N} \sum d_i$$

Both approaches normalize to [0,1] via $SA$ transformation, but only cosine provides scale-invariant comparison.

### A.3 Error Propagation

Standard error of $SA$ metric computed via bootstrap resampling (1000 iterations). For each iteration:

1. Sample translation pairs with replacement
2. Compute $SA$ score
3. Calculate standard deviation across bootstrap samples

Reported as: $SA = \mu \pm SEM$

### A.4 Linguistic Grounding: Inter/Intra Terminology

Intra-lingual (within a single language) measures semantic diversity within each language's vocabulary (normalization baseline).

Inter-lingual (between different languages) measures whether translations of the same concept cluster together across language boundaries.

This terminology directly describes what's being measured, providing immediate interpretability: inter/intra ratio immediately conveys cross-language comparison to within-language baseline.

### A.5 Why $SA$ (Cosine) is Superior to $SA$ (Euclidean)

Empirical Evidence: We discovered Euclidean $SA$ shows dataset-dependent inconsistency for the same model, while Cosine $SA$ remains stable:

Case Study: Qwen3-0.6B Model

| Dataset | $SA$ (Euclidean) | $SA$ (Cosine) |
|---|---|---|
| DS1 (Peterg, 349w) | 0.456 | 0.490 |
| DS2 (Cultural, 341w) | 0.702 | 0.520 |
| Volatility ($\Delta$) | 0.246 | 0.030 |

Analysis:

- Euclidean inconsistency: Same model yields $SA = 0.46$ (poor) on DS1 but $SA = 0.70$ (excellent!) on DS2—contradictory assessments across datasets

- Cosine consistency: Same model yields $SA \approx$ 0.49-0.52 (borderline) on both datasets—correctly identifies model as barely passing threshold

- 8× higher volatility: Euclidean shows $\Delta = 0.246$ swing vs Cosine $\Delta = 0.030$. From PHATE charts, we see embedding structure is abnormal.

Root Cause: Euclidean distances are scale-dependent:

- DS1: Intra=53.9, Inter=64.4 (magnitudes 50-60)

- DS2: Intra=51.4, Inter=21.8 (magnitudes 20-50)

- Different absolute embedding scales → incomparable ratios

Cosine Advantage: Angular distance [0,2], invariant to magnitude:

- DS1: Intra 0.19, Inter 0.20 (normalized scale)

- DS2: Intra 0.22, Inter 0.20 (normalized scale)

- Same normalized scale → directly comparable ratios

Conclusion: Cosine $SA$ enables fair cross-dataset comparison and absolute quality assessment. Euclidean $SA$ conflates embedding scale with alignment quality, yielding unreliable benchmarking. Therefore, we adopt Cosine as primary metric.

## A.6 Practical Diagnostic Workflow Using $SA$ + PHATE

We recommend combining quantitative $SA$ metrics with qualitative PHATE visualization for comprehensive model evaluation. This two-stage protocol uses Cosine $SA$ as primary assessment and Euclidean spreads as diagnostic indicators.

Stage 1: Primary Quality Assessment (Cosine $SA$)

Compute $SA$ using cosine distance (magnitude-invariant) and interpret using standardized thresholds:

- $SA \geq 0.60$: Great alignment (Tier 1) → Deploy with confidence

- $0.50 \leq SA < 0.60$: Good alignment (Tier 2) → Task-dependent suitability

- $SA < 0.50$: Non-alignment (Tier 3) → Proceed to Stage 2 for diagnosis

Stage 2: Diagnostic Analysis (Euclidean Spreads + PHATE)

For models with $SA < 0.60$ (good or non-aligned), examine Euclidean spreads and PHATE visualization to diagnose embedding space pathologies:

1. Check absolute magnitudes:
   - High magnitudes (Inter/Intra > 50): Poor embedding normalization
   - Near-zero magnitudes (Inter/Intra < 1): Potential collapse
   - Moderate magnitudes (1-20): Healthy embedding space

2. Check cross-dataset variance:
   - High variance (>3× across datasets): Inconsistent embedding quality
   - Low variance (<2× across datasets): Stable embeddings

3. Visual validation with PHATE:

- Language clustering (separate spatial regions) → Alignment failure
- Mixed clusters (interleaved points) → Partial alignment
- Translation co-location (overlapping clusters) → Good alignment

Case Study: XLM-RoBERTa model

Figure E.6 shows PHATE plots for XLM-R on 4 datasets. Applying the diagnostic workflow:

Stage 1 - Primary Assessment:

- $SA_{\cos} = 0.433 \pm 0.036$ (DS1)

- Decision: $SA < 0.50$ → Poor alignment (Tier 3) → Proceed to Stage 2

Stage 2 - Diagnostic Analysis:

Euclidean spreads (DS1):

- Inter = 0.106, Intra = 0.095 (magnitudes < 1)

- Diagnosis 1: Near-collapsed embedding space (magnitudes ≪ 1)

- Cross-dataset check: DS1 (0.106), DS2 (0.105), DS3 (0.109), DS4 (0.098)

- Diagnosis 2: Consistent collapse across all datasets (low variance)

Cosine spreads (DS1):

- Inter = 0.0060, Intra = 0.0046

- Inter/Intra ratio = 1.30 → Translations are farther than random pairs

- Diagnosis 3: Active language separation (Inter > Intra)

PHATE visualization (DS1):

- Chinese (red) clustered entirely on LEFT side

- English (blue) clustered entirely on RIGHT side

- Virtually NO overlap between language clusters

- Diagnosis 4: Visual confirmation of very poor semantic alignment

Final Diagnosis:

XLM-RoBERTa exhibits alignment failure despite MLM training on 100 languages. The model learns language-specific patterns (causing spatial separation) without cross-lingual semantic binding. The near-collapsed embedding magnitudes ($<1$) combined with active language separation (Inter>Intra) indicate that unsupervised multilingual pre-training is fundamentally insufficient for cross-lingual alignment.

Recommendation: Avoid XLM-RoBERTa for applications requiring genuine cross-lingual understanding. Use models with explicit translation-pair supervision (LaBSE, S-BERT, USE) instead.

Contrast with Successful Model (LaBSE on DS1):

- $SA_{cos} = 0.807 \pm 0.005$ (Excellent, Tier 1)

- Euclidean: Inter $= 6.20$, Intra $= 12.17$ (healthy magnitudes)

- Cosine: Inter $= 0.090$, Intra $= 0.379$ (Inter is $4\times$ smaller!)

- PHATE: Complete language interleaving (red/blue points mixed)

- Translation-pair supervision $\rightarrow$ Strong cross-lingual binding

This diagnostic workflow demonstrates how combining Cosine SA (primary), Euclidean spreads (diagnostic), and PHATE visualization (visual validation) provides complete assessment of embedding model quality for cross-lingual tasks.

## Appendix B: Semanscope Implementation

Semanscope tool will be released as open-source. It combines PHATE manifold learning with $SA$ metric to test the translation-equivalence assumption underlying multilingual embeddings. The tool is designed to be user-friendly and accessible, with a focus on providing clear and actionable insights into the performance of multilingual embedding models.

### B.1 System Architecture

- Frontend: Streamlit web application

- Backend: Python with NumPy, scikit-learn, PHATE library

- Visualization: Plotly (interactive mode), ECharts (publication mode with PDF export)

- Caching: Joblib for embedding persistence (per-word granular)

- Parallelization: loky backend for true multiprocessing (avoids Python GIL)

### B.2 PHATE Configuration

- knn: 15 (optimal for datasets 1000-4000 samples)

- decay: 40 (controls diffusion process)

- t: auto (automatic optimal diffusion time)

- gamma: 1 (information distance parameter)

- n_components: 2 (2D visualization)

### B.3 PHATE Preprocessing Pipeline

1. StandardScaler normalization: Zero mean, unit variance per dimension

2. Epsilon noise: Add 1e-10 for numerical stability

3. Duplicate filtering: Remove duplicate embeddings per language (prevents artificial clustering)

4. Unique word extraction: Flatten multi-meaning "|" delimiters, deduplicate via set()

### B.4 Key Optimizations

1. Per-Word Granular Caching ($30\times$ efficiency):

- Challenge: Multi-meaning words like "十 | 完整" (ten|complete) expand to 4 pairs when combined with "ten|complete"

- Solution: Cache embeddings at word level, not phrase level

- Result: 9,397 $\rightarrow$ 327 unique embeddings/language

2. True Multiprocessing ($4.6\times$ speedup):

- Challenge: Python GIL limits threading performance

- Solution: loky backend for process-based parallelism

- Result: Linear scaling up to $N$ CPU cores

3. Automatic $SA$ Legend Generation:

- Compute $SA$ metrics from inter/intra spreads (Cosine primary, Euclidean supplementary)

- Format as compact 2-line text

- Position at optimal location (top-right 9%/9%)

- Apply to both Plotly and ECharts backends

- Export in publication-ready PDF/PNG

### B.5 Usage Example

from semanscope import SemanticAffinityAnalyzer

```
analyzer = SemanticAffinityAnalyzer(
    model_name="LaBSE",
    dataset="ICML-1-peterg",
    languages=["chn", "enu"]
)

results = analyzer.compute_semantic_affinity()
print(f"SA_cosine = {results['cosine'].score:.4f}")

analyzer.plot_phate(
    output_format="pdf",
    output_path="labse_phate_with_legend.pdf",
    include_sa_legend=True  # Automatic overlay
)
```

## Appendix C: Experimental Protocol

### C.1 Dataset Specifications

Table C.1 lists the four datasets (DS1/DS2/DS3/DS4) used in this study. Each dataset contains translation pairs across English and Chinese, with varying semantic domains and difficulty levels. DS1-DS3 are machine translated from English to Chinese, DS4 is machine translated from Chinese to English, then manually reviewed and revised. All datasets will be released as part of Semanscope open-source git repository upon publication.

Here these datasets serve research and illustrative purposes only. For production model selection, practitioners should construct domain-specific datasets using vocabulary relevant to their actual use cases. Domain-specific semantic alignment may differ significantly from general-purpose benchmarks—a model performing well on these research datasets may show different $SA$ scores on specialized terminology (e.g., medical oncology terms, quantitative finance jargon, legal contract language). Semanscope's open-source framework enables practitioners to evaluate models on their own translation pairs, ensuring the selected model aligns well on deployment-specific vocabulary rather than general concepts.

### C.2 Model Specifications

Table C.2 lists the 13 models evaluated in this study, covering a range of architectures (BERT, Transformer, LLM), sizes (335M to 8B parameters), and training objectives (translation-pair supervision, semantic similarity, next-token prediction, MLM-only). This diverse selection enables comprehensive assessment of how architecture, scale, and training objective impact cross-lingual semantic alignment. Model column shows alias used throughout paper. Description shows full model name.

### C.3 Model Loading and Embedding Generation

HuggingFace and Local Models (5):

- LaBSE: sentence-transformers

- Universal Sentence Encoder: TensorFlow Hub

- Sentence-BERT: sentence-transformers

- E5-Large-Instruct: sentence-transformers

- Qwen3-0.6B: sentence-transformers

- mBERT, XLM-R: sentence-transformers

API Models (8):

- Qwen3-4B, Qwen3-8B: OpenRouter API

- Gemini-001: OpenRouter API

- OpenAI ada-002, 3-small, 3-large: OpenRouter API

Embedding Cache: Per-word granular caching. Each word embedded once, cached, reused across experiments. Reduces $9{,}397 \rightarrow 327$ embeddings/language ($30\times$ efficiency) and API costs.

### C.4 $SA$ Computation Pipeline

For each model × dataset × language pair:

1. Load dataset text files (e.g., ICML-1-peterg-chn.txt, ICML-1-peterg-enu.txt)

2. Generate embeddings (use cache if available)

3. Compute intra-lingual spread (pairwise distances within each language)

4. Compute inter-lingual spread (pairwise distances across translation pairs, handling multi-meaning expansions)

| Key | Dataset | Description | Words |
|-----|---------|-------------|-------|
| DS1 | ICML-1-peterg | Gärdenfors conceptual space words (colors, shapes, animals, spatial relations) | 349 |
| DS2 | ICML-2-cultural-nuance | Culturally-sensitive concepts (honor, kinship terms, philosophical concepts) | 190 |
| DS3 | ICML-3-challenge | Linguistic edge cases (untranslatable words, polysemes, false friends) | 86 |
| DS4 | ICML-4-zinets | Oracle Bone Script characters (1200 BCE) representing universal primitives (sun, water, hand, eye) | 327 |

Table C.1. Datasets Used in This Study

| Model | Training Objective | Description |
|-------|-------------------|-------------|
| LaBSE | Translation pairs (17B sentences) | LaBSE |
| USE | Semantic similarity (multi-task) | Universal-Sentence-Encoder-Multilingual |
| Sentence-BERT | Multilingual NLI | Sentence-BERT Multilingual |
| Gemini-001 | General purpose | Gemini-Embedding-001 (OpenRouter) |
| Qwen3-8B | Next-token prediction | Qwen3-Embedding-8B (OpenRouter) |
| Qwen3-4B | Next-token prediction | Qwen3-Embedding-4B (OpenRouter) |
| Qwen3-0.6B | Next-token prediction | Qwen3-Embedding-0.6B |
| OpenAI-3-Large | General purpose | OpenAI Text-Embedding-3-Large (OpenRouter) |
| OpenAI-3-Small | General purpose | OpenAI Text-Embedding-3-Small (OpenRouter) |
| OpenAI-Ada-002 | General purpose | OpenAI Text-Embedding-Ada-002 (OpenRouter) |
| E5-Large | Instruction tuning | Multilingual-E5-Large-Instruct-v2 |
| mBERT | MLM only (104 languages) | mBERT |
| XLM-R | MLM only (100 languages) | XLM-RoBERTa-v2 |

Table C.2. Models Evaluated in This Study

5. Calculate $SA$ metrics (Cosine primary, Euclidean supplementary)

6. Bootstrap resampling (1000 iterations) for error estimation

7. Generate PHATE manifold with automatic $SA$ legend overlay

8. Export results (CSV, JSON) and visualization charts (PDF, PNG)

Parallelization: multi-core CPU with loky backend for true multiprocessing. Avoids Python GIL limitations.

### C.5 Quality Control

Collapse Detection: Flag models with max embedding distance < 1e-6 (indicates degenerate embeddings)

Validation: Cross-check $SA$ scores with visual inspection of PHATE plots. Models with $SA > 0.60$ should show clear language interleaving; models with $SA < 0.50$ should show language separation and poor semantic alignment.

## Appendix D: Experimental Results

### Key Findings From Tables D.1-D.2:

1. Three-tier structure (Cosine): Top BERT ($SA \geq 0.60$), LLM Plateau ($0.50 \leq SA < 0.60$), Failures ($SA < 0.50$);

2. Dataset difficulty gradient: DS1/DS2 > DS3 > DS4 across all models;

3. Top BERT steeper drops (-25-31%) vs LLM robustness (-15-17%);

4. E5-Large in LLM tier despite BERT architecture;

5. LaBSE most robust Tier 1 model;

6. Cosine (primary) provides stable cross-dataset ranking; Euclidean (supplementary) shows dataset-dependent volatility;

7. Critical finding persists: translation-pair training essential.

| Model | DS1 [349] | DS2 [341] | DS3 [177] | DS4 [769] | Avg $SA$ |
|---|---|---|---|---|---|
| LaBSE | $0.807 \pm 0.005$ | $0.804 \pm 0.005$ | $0.575 \pm 0.011$ | $0.583 \pm 0.021$ | 0.692 |
| Sentence-BERT | $0.762 \pm 0.040$ | $0.799 \pm 0.029$ | $0.583 \pm 0.047$ | $0.540 \pm 0.108$ | 0.671 |
| USE | $0.767 \pm 0.007$ | $0.789 \pm 0.008$ | $0.575 \pm 0.012$ | $0.545 \pm 0.055$ | 0.669 |
| OpenAI-3-Large | $0.644 \pm 0.002$ | $0.660 \pm 0.004$ | $0.578 \pm 0.005$ | $0.545 \pm 0.013$ | 0.607 |
| OpenAI-3-Small | $0.637 \pm 0.003$ | $0.643 \pm 0.005$ | $0.554 \pm 0.008$ | $0.550 \pm 0.013$ | 0.596 |
| Gemini-001 | $0.618 \pm 0.010$ | $0.624 \pm 0.011$ | $0.574 \pm 0.014$ | $0.532 \pm 0.020$ | 0.587 |
| Qwen3-8B | $0.595 \pm 0.030$ | $0.619 \pm 0.021$ | $0.537 \pm 0.025$ | $0.503 \pm 0.002$ | 0.564 |
| Qwen3-4B | $0.594 \pm 0.031$ | $0.609 \pm 0.024$ | $0.517 \pm 0.022$ | $0.510 \pm 0.007$ | 0.558 |
| E5-Large | $0.594 \pm 0.006$ | $0.601 \pm 0.003$ | $0.513 \pm 0.008$ | $0.477 \pm 0.012$ | 0.546 |
| OpenAI-Ada-002 | $0.589 \pm 0.004$ | $0.586 \pm 0.003$ | $0.520 \pm 0.009$ | $0.503 \pm 0.013$ | 0.550 |
| mBERT | $0.507 \pm 0.016$ | $0.517 \pm 0.011$ | $0.482 \pm 0.007$ | $0.473 \pm 0.006$ | 0.495 |
| Qwen3-0.6B | $0.490 \pm 0.022$ | $0.519 \pm 0.009$ | $0.482 \pm 0.006$ | $0.427 \pm 0.004$ | 0.480 |
| XLM-R | $0.433 \pm 0.035$ | $0.443 \pm 0.049$ | $0.472 \pm 0.008$ | $0.446 \pm 0.068$ | 0.449 |

Table D.1. Semantic Affinity Scores (Cosine Distance - PRIMARY) - Complete Results. Dataset keys (DS1-DS4) defined in Table C.1.

| Model | DS1 [349] | DS2 [341] | DS3 [177] | DS4 [769] | Avg $SA$ |
|---|---|---|---|---|---|
| LaBSE | $0.692 \pm 0.004$ | $0.679 \pm 0.003$ | $0.551 \pm 0.005$ | $0.560 \pm 0.014$ | 0.620 |
| USE | $0.672 \pm 0.006$ | $0.679 \pm 0.005$ | $0.548 \pm 0.006$ | $0.530 \pm 0.029$ | 0.607 |
| Sentence-BERT | $0.668 \pm 0.023$ | $0.686 \pm 0.019$ | $0.554 \pm 0.024$ | $0.522 \pm 0.058$ | 0.607 |
| OpenAI-3-Large | $0.582 \pm 0.002$ | $0.585 \pm 0.002$ | $0.545 \pm 0.003$ | $0.525 \pm 0.007$ | 0.559 |
| OpenAI-3-Small | $0.577 \pm 0.002$ | $0.576 \pm 0.003$ | $0.532 \pm 0.004$ | $0.528 \pm 0.007$ | 0.553 |
| Gemini-001 | $0.567 \pm 0.005$ | $0.566 \pm 0.005$ | $0.543 \pm 0.008$ | $0.517 \pm 0.010$ | 0.548 |
| Qwen3-8B | $0.557 \pm 0.016$ | $0.564 \pm 0.012$ | $0.525 \pm 0.012$ | $0.505 \pm 0.002$ | 0.538 |
| Qwen3-4B | $0.556 \pm 0.016$ | $0.558 \pm 0.013$ | $0.517 \pm 0.011$ | $0.507 \pm 0.003$ | 0.535 |
| OpenAI-Ada-002 | $0.551 \pm 0.002$ | $0.546 \pm 0.002$ | $0.515 \pm 0.005$ | $0.504 \pm 0.007$ | 0.529 |
| E5-Large | $0.555 \pm 0.003$ | $0.554 \pm 0.002$ | $0.512 \pm 0.004$ | $0.490 \pm 0.006$ | 0.528 |
| mBERT | $0.507 \pm 0.008$ | $0.511 \pm 0.006$ | $0.492 \pm 0.004$ | $0.487 \pm 0.003$ | 0.499 |
| Qwen3-0.6B | $0.502 \pm 0.011$ | $0.512 \pm 0.005$ | $0.496 \pm 0.003$ | $0.469 \pm 0.002$ | 0.495 |
| XLM-R | $0.475 \pm 0.018$ | $0.478 \pm 0.025$ | $0.498 \pm 0.004$ | $0.477 \pm 0.035$ | 0.482 |

Table D.2. Semantic Affinity Scores (Euclidean Distance - SUPPLEMENTARY) - Complete Results. Dataset keys (DS1-DS4) defined in Table C.1. Note: Euclidean scores show dataset-dependent volatility (see Appendix A.5).

## Appendix E: PHATE Visualizations

This appendix provides more PHATE manifold-learning charts for 6 models across 4 datasets, organized in 2×2 grids showing varying performance across semantic domains. All $SA$ scores shown are cosine distance (primary metric).

- E.1 LaBSE (Tier 1: $SA = 0.70$, Translation-Pair Supervision)

- E.2 OpenAI-3-Large (Tier 2: $SA = 0.61$, LLM Plateau)

- E.3 E5-Large-Instruct (Tier 2: $SA = 0.55$, LLM Plateau)

- E.4 Qwen3-0.6B (Tier 3: $SA = 0.48$, Failure)

- E.5 mBERT (Tier 3: $SA = 0.50$, MLM-Only Failure)

- E.6 XLM-RoBERTa-v2 (Tier 3: $SA = 0.45$, Failure)

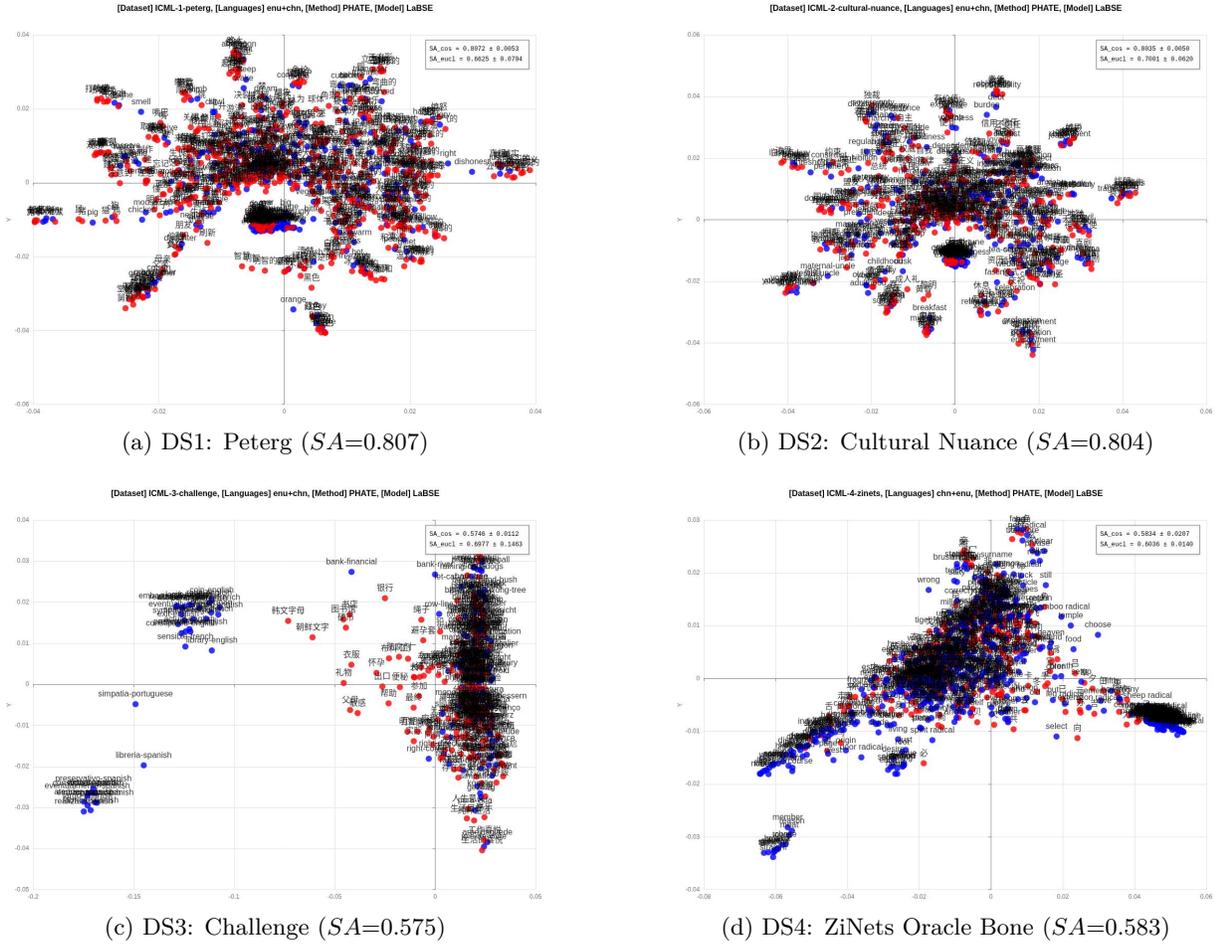

(a) DS1: Peterg ($SA$=0.807)

(b) DS2: Cultural Nuance ($SA$=0.804)

(c) DS3: Challenge ($SA$=0.575)

(d) DS4: ZiNets Oracle Bone ($SA$=0.583)

Figure E.1. LaBSE: Beautiful language interleaving across all datasets. Best overall alignment with translation-pair supervision.

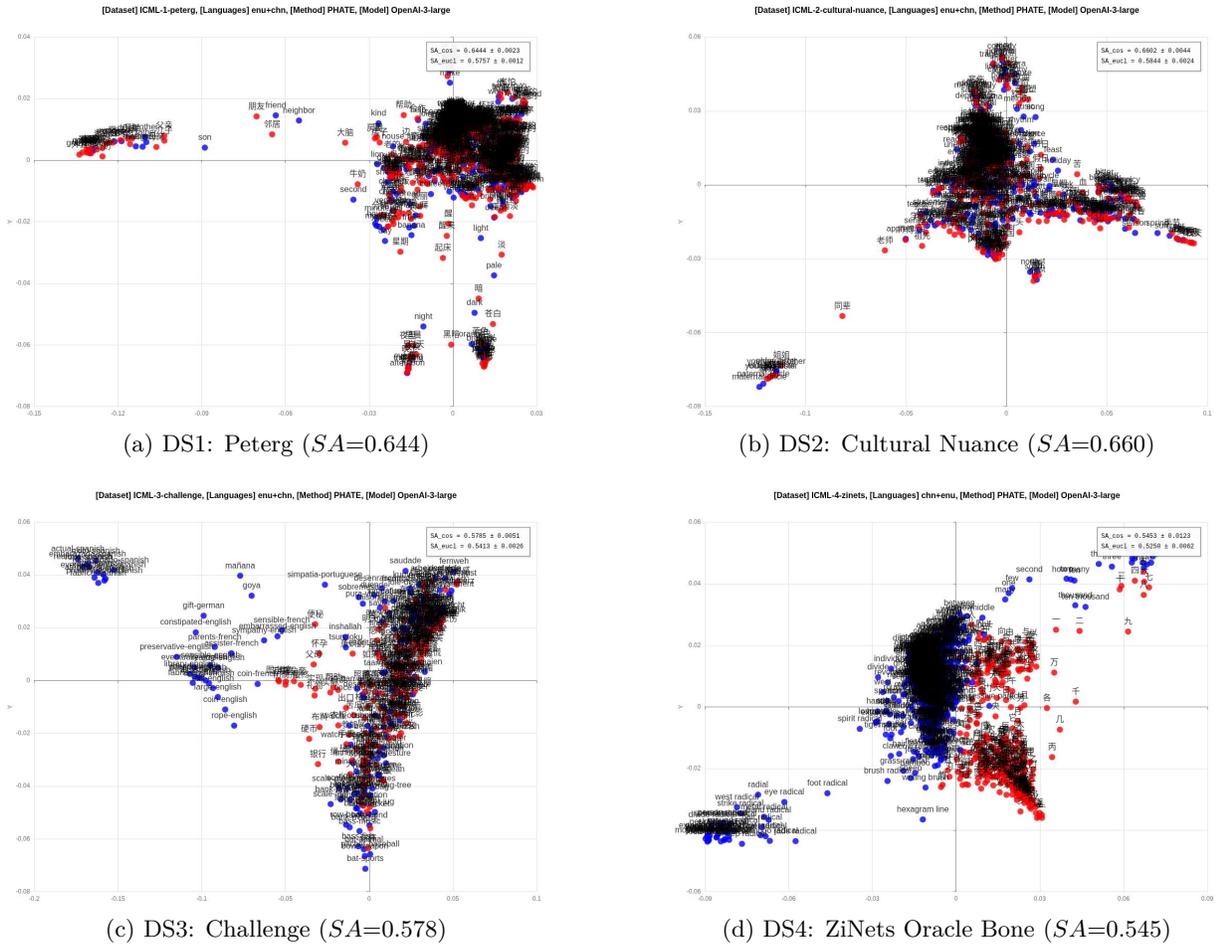

(a) DS1: Peterg ($SA$=0.644)

(b) DS2: Cultural Nuance ($SA$=0.660)

(c) DS3: Challenge ($SA$=0.578)

(d) DS4: ZiNets Oracle Bone ($SA$=0.545)

Figure E.2. OpenAI-3-Large: Moderate cross-lingual mixing with visible clustering. Robust across domains (-17% drop DS1→DS4).

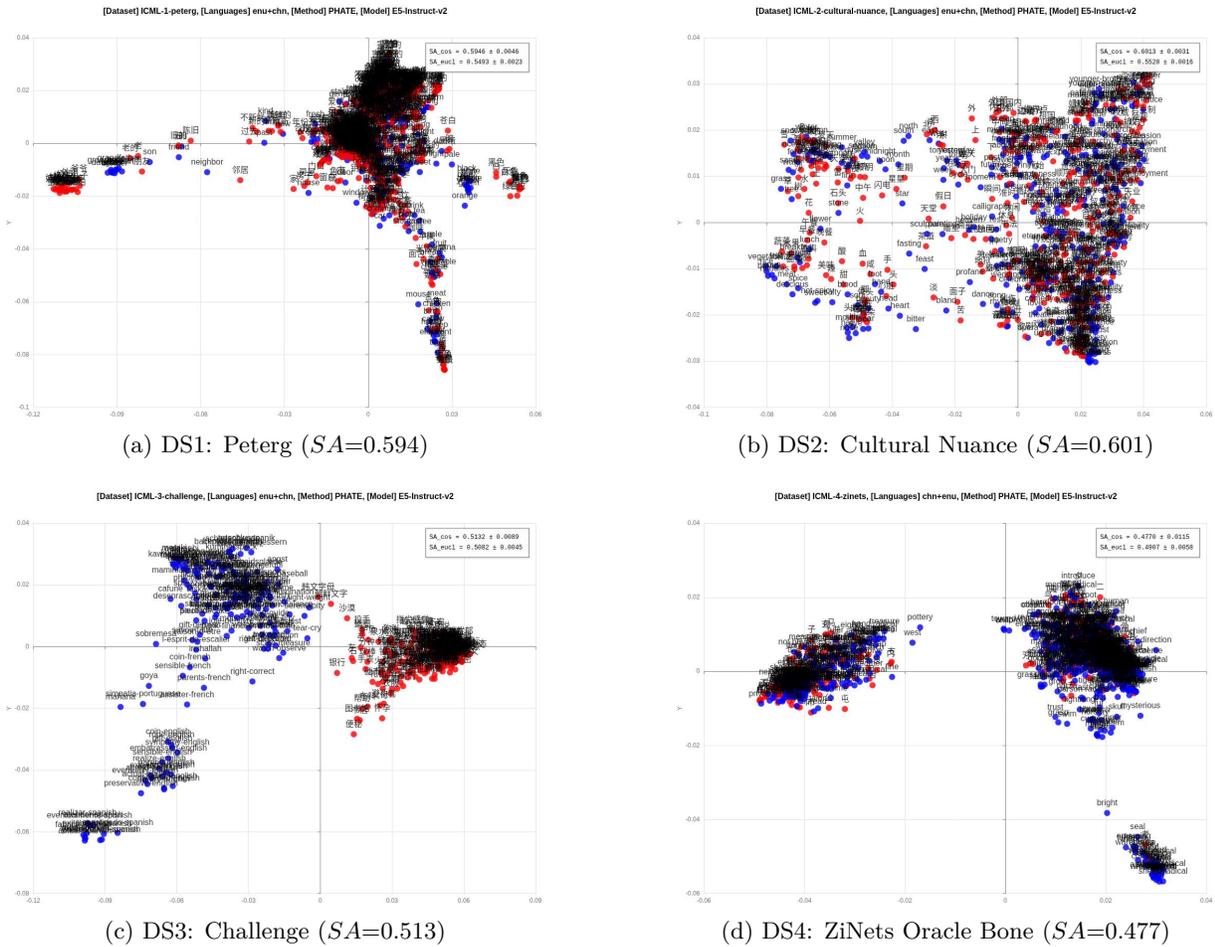

(a) DS1: Peterg ($SA$=0.594)

(b) DS2: Cultural Nuance ($SA$=0.601)

(c) DS3: Challenge ($SA$=0.513)

(d) DS4: ZiNets Oracle Bone ($SA$=0.477)

Figure E.3. E5-Large-Instruct: Moderate cross-lingual mixing with visible clustering.

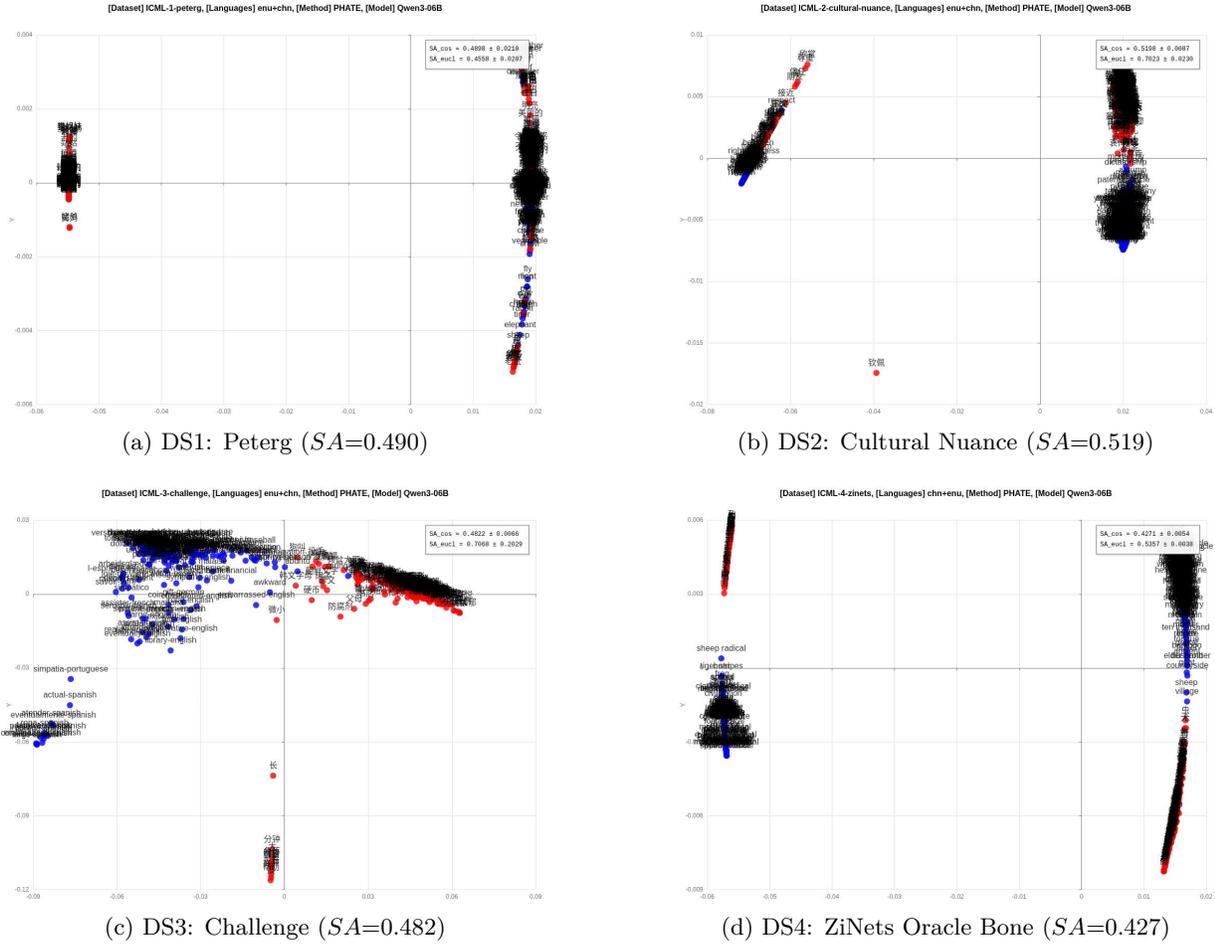

(a) DS1: Peterg ($SA$=0.490)

(b) DS2: Cultural Nuance ($SA$=0.519)

(c) DS3: Challenge ($SA$=0.482)

(d) DS4: ZiNets Oracle Bone ($SA$=0.427)

Figure E.4. Qwen3-0.6B: Partial language separation visible. Fails alignment threshold ($SA < 0.50$). Next-token prediction insufficient for cross-lingual binding.

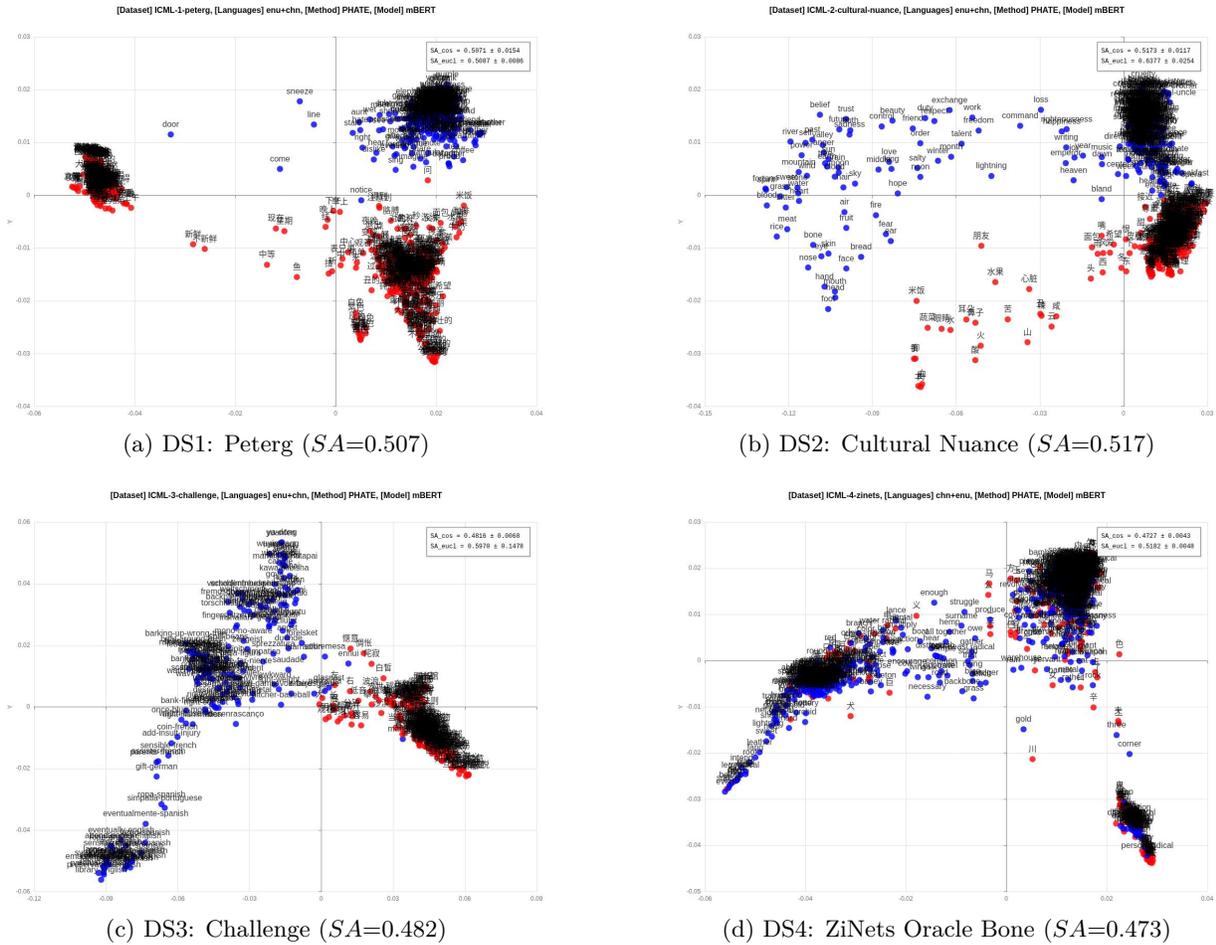

(a) DS1: Peterg ($SA$=0.507)

(b) DS2: Cultural Nuance ($SA$=0.517)

(c) DS3: Challenge ($SA$=0.482)

(d) DS4: ZiNets Oracle Bone ($SA$=0.473)

Figure E.5. mBERT: Significant language separation despite 104-language MLM training. MLM-only objective learns language-specific patterns.

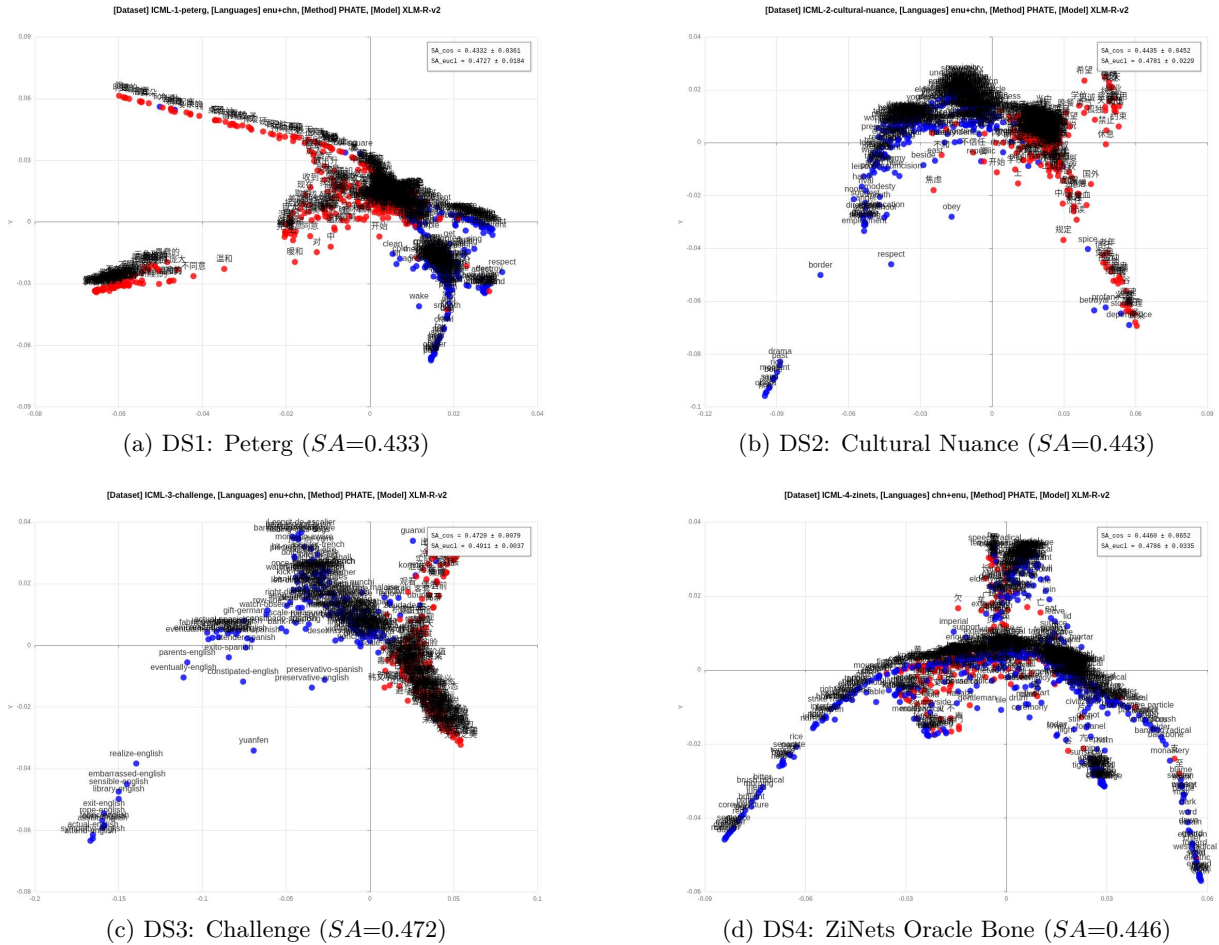

(a) DS1: Peterg ($SA$=0.433)

(b) DS2: Cultural Nuance ($SA$=0.443)

(c) DS3: Challenge ($SA$=0.472)

(d) DS4: ZiNets Oracle Bone ($SA$=0.446)

Figure E.6. **XLM-RoBERTa-v2:** Significant language separation despite multi-language MLM training. MLM-only objective learns language-specific patterns.